\documentclass[conference]{IEEEtran}
\IEEEoverridecommandlockouts
\usepackage{multirow}
\usepackage{tabularx}
\usepackage{makecell}
\usepackage{stfloats}
\usepackage{cite}
\usepackage[numbers]{natbib}
\usepackage{amsmath,amssymb,amsfonts}
\usepackage{graphicx}
\usepackage{textcomp}
\usepackage{xcolor}
\usepackage{algorithm}
\usepackage{algorithmic}
\usepackage{hyperref}

\def\BibTeX{{\rm B\kern-.05em{\sc i\kern-.025em b}\kern-.08em
    T\kern-.1667em\lower.7ex\hbox{E}\kern-.125emX}}

\begin{document}
\title{OC\textsuperscript{4}-ReID: Occluded Cloth-Changing Person Re-Identification}
\author{
Zhihao Chen$^{1}$ \quad Yiyuan Ge$^{2*}$ \quad  Yanyan Lv $^{3}$ \quad Ziyang Wang$^{4*}$ \quad  \quad Mingya Zhang$^{5}$\\
    $^{1}$ Beijing University of Posts and Telecommunications \quad $^{2}$ South China University of Technology \quad \\
    $^{3}$ Nanjing University of Information Science \& Technology \quad $^{4}$ University of Oxford \quad
    $^{5}$ Nanjing University \\
    zhihaochen666@bupt.edu.cn \quad 202510182871@mail.scut.edu.cn \quad yylv@nuist.edu.cn \quad \\ ziyang.wang17@gmail.com \quad \quad dg20330034@smail.nju.edu.cn
}

\maketitle

\begin{abstract}
The study of Cloth-Changing Person Re-identification (CC-ReID) focuses on retrieving specific pedestrians when their clothing has changed, typically under the assumption that the entire pedestrian images are visible. Pedestrian images in real-world scenarios, however, are often partially obscured by obstacles, presenting a significant challenge to existing CC-ReID systems. In this paper, we introduce a more challenging task termed Occluded Cloth-Changing Person Re-identification (OC\textsuperscript{4}-ReID), which simultaneously addresses two challenges of clothing changes and occlusions. Concretely, we construct two new datasets, Occ-LTCC and Occ-PRCC, based on original CC-ReID datasets to include random occlusions of key pedestrians components (e.g., head, torso). Moreover, a novel benchmark is proposed for OC\textsuperscript{4}-ReID incorporating a Train-Test Micro Granularity Screening (T\textsuperscript{2}MGS) module to mitigate the influence of occlusions and proposing a Part-Robust Triplet (PRT) loss for partial features learning. Comprehensive experiments on the proposed datasets, as well as on two CC-ReID benchmark datasets demonstrate the superior performance of proposed method against other state-of-the-art methods. The codes and datasets are available at: \textcolor{red}{https://github.com/1024AILab/OC4-ReID}.
\end{abstract}

\begin{IEEEkeywords}
Person Re-identification, Cloth-Changing, Occlusion.
\end{IEEEkeywords}

\section{Introduction}
Cloth-Changing Person Re-identification (CC-ReID) aims to retrieve target pedestrians across instances where their clothing has changed \cite{b1, b2, b3}. To mitigate clothing bias, existing methods utilize deep neural networks to learn features that are less dependent on clothing, often assuming that the pedestrian images are minimally occluded \cite{b4, b5, b1, b7, b8}. In real-world application scenarios, pedestrian images captured by practical cameras, however, are often occluded. This occlusion reduces available feature information and can cause the model to make incorrect matches. 

\begin{figure}[t]  
\centering  
\includegraphics[width=\linewidth]{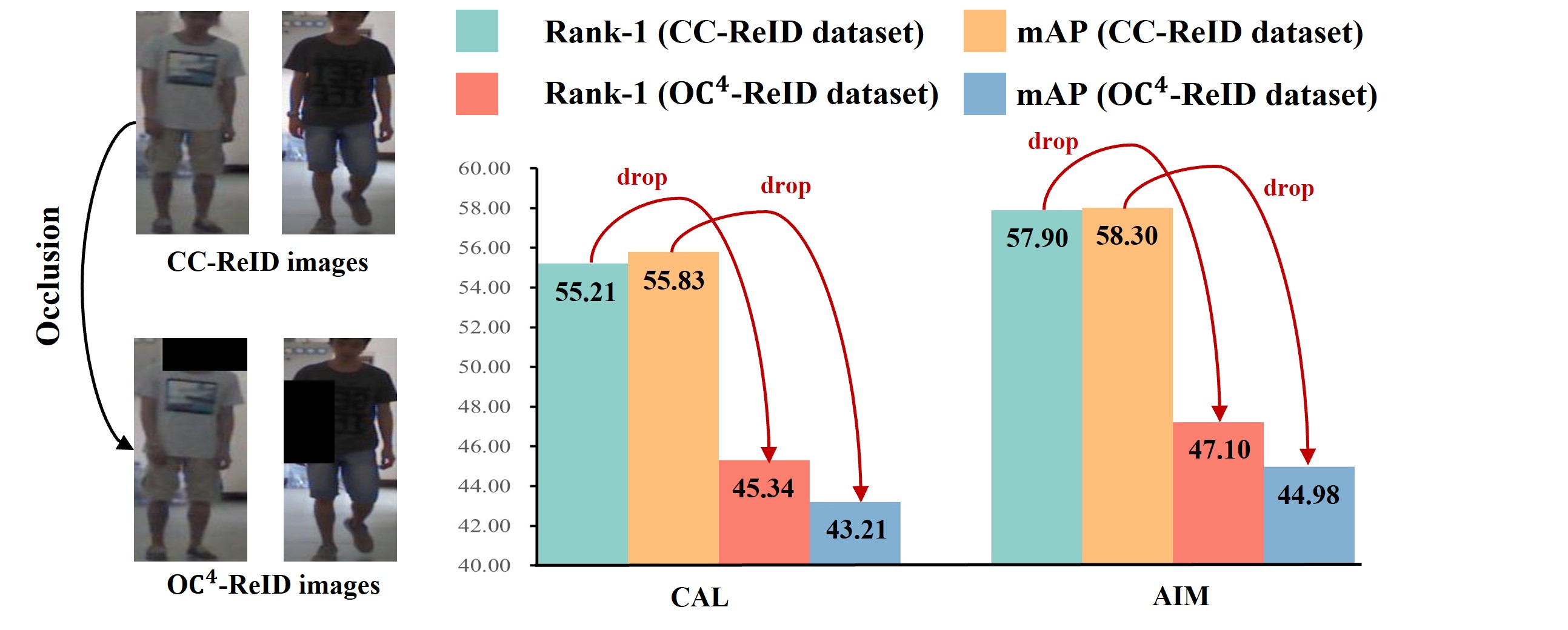}
\caption{We construct OC\textsuperscript{4}-ReID dataset based on the CC-ReID dataset by introducing random occlusions on various human body parts (e.g., head, arms) . Fair training and testing results indicate that traditional CC-ReID methods (e.g., CAL \cite{b9} and AIM \cite{b10}) perform worse on the OC\textsuperscript{4}-ReID dataset.}
\label{fig1}
\end{figure}

Fig. \ref{fig1} illustrates the evaluation of classical CC-ReID architectures \cite{b9, b10} on both the standard CC-ReID dataset and our newly constructed OC\textsuperscript{4}-ReID dataset, where parts of the pedestrian's body are randomly occluded. Our experimental results indicate that occlusions into the CC-ReID dataset significantly impairs the model's performance, with CAL \cite{b9} and AIM \cite{b10} reducing in Rank-1 by 9.87\% and 10.80\%, and in mAP by 12.62\% and 13.32\%, respectively. 

Given the significant impact of occlusions on CC-ReID, we introduce a novel and more challenging task, termed \textbf{Occ}luded \textbf{C}loth-\textbf{C}hanging Person \textbf{Re-id}entification (OC\textsuperscript{4}-ReID). This task aims to identify pedestrians while overcoming the dual challenges of occlusions and clothing changes.
Particularly, OC\textsuperscript{4}-ReID is critical for suspects tracking, as suspects often change clothes and utilize obstacles to hide their bodies during escape, which misleads conventional ReID tracking systems. Compared to CC-ReID, OC\textsuperscript{4}-ReID poses two primary challenges: (i) occlusion leads to even less cloth-unrelated pedestrian discrimination information; (ii) blurred and low-quality occlusion regions further interfere with effective features extraction. 

In this paper, we propose two specialized OC\textsuperscript{4}-ReID datasets and a simple yet effective benchmark for OC\textsuperscript{4}-ReID task. \textbf{Firstly}, We adopt a pre-trained human parser \cite{b11} to locate the body components of pedestrians in PRCC \cite{b12} and LTCC \cite{b13} datasets, then randomly select one component to occlude, constructing the OC\textsuperscript{4}-ReID datasets. 
\textbf{Secondly}, to mitigate the impact of occlusion and mine discriminative local representations, we propose a \textbf{T}rain-\textbf{T}est \textbf{M}icro \textbf{G}ranularity \textbf{S}creening (T\textsuperscript{2}MGS) module, in which a quality predictor is maintained to evaluate each micro granularity features. During testing, low-quality features are discarded to suppress occlusions. \textbf{Thirdly}, we also propose a \textbf{P}art-\textbf{R}obust \textbf{T}riplet loss (PRT) to supervise partial features, which aims to combine all the body parts and mine truly discriminative representations. In addition, PRT is robust to clothing features in pedestrians' partial appearances and emphasises the extraction of cloth-unrelated identity information. 
The main contributions of this paper are summarized as follows:
\begin{itemize}
\item To the best of our knowledge, the OC\textsuperscript{4}-ReID task is introduced for the first time, and we construct two specialized datasets Occ-PRCC and Occ-LTCC.
\item We propose a novel benchmark for OC\textsuperscript{4}-ReID, introducing the Train-Test Micro Granularity Screening (T\textsuperscript{2}MGS) module to against occlusion.
\item We introduce a Part-Robust Triplet (PRT) loss to supervise the learning of body partial features in pedestrians, making model robust to clothing and occlusion regions.
\item The experimental results in both cloth-changing and occluded cloth-changing scenarios demonstrate the effectiveness of the proposed method.
\end{itemize}

\section{Related Works}
\subsection{Occluded Person Re-identification (OCC-ReID)}

Person ReID matches the target pedestrian from the image gallery, crossing different scenarios and views  \cite{b28, b29, b30, b31}. All of the above ReID methods assume that the entire body of the pedestrian is visible, ignoring the more challenging case of occlusion. Mainstream OCC-ReID methods utilize pose estimation \cite{b36} and body parsing \cite{b35, b37} to localize body parts, aiming to mitigate the effects of occlusion. Wang et al. \cite{b36} used labeled poses to learn visible local features as well as topological information during the training and testing phases, achieving high detection accuracy. Song et al. \cite{b35} utilized human parsing masks to generate attention maps, enabling the extraction of discriminative and robust features in occluded scenarios. However, these methods rely on maintaining an offline pose or segmentation encoder, which limits model flexibility and incurs significant computational overhead.


\subsection{Cloth-changing Person Re-identification (CC-ReID)}
CC-ReID considers a long-term retrieval situation where the pedestrian's clothing changes. To tackle this, the researchers employed different approaches including GAN \cite{b1}, 3D information \cite{b40}, human contour \cite{b3}, and adversarial loss \cite{b9}. Xu et al. \cite{b1} enhanced the model's ability to learn cloth-unrelated features by reconstructing intra-class examples to minimize feature differences while generating adversarial dressing examples across classes. Chen et al. \cite{b40} introduced 3D Shape Learning (3DSL), leveraging 3D body reconstruction as an auxiliary task to extract texture-insensitive 3D shape features for CC-ReID. Cui et al. \cite{b3} achieved controlled reconstruction by introducing the cloth-unrelated, cloth-related, and pedestrian contour features. Gu et al. \cite{b9} introduced a cloth-based adversarial loss which first trains a cloth classifier to capture cloth-related features, then shifts focus to learning cloth-unrelated features in the second stage. 

Although the above methods \cite{b1, b3, b9, b40} mitigate clothing variations to a certain extent, they are less effective when additional occlusion disturbances are introduced. Therefore, the aim of this paper is to challenge an even tougher scenario with integrated cloth changes and occlusions.

\section{OC\textsuperscript{4}-ReID Dataset}
\begin{figure}[!t]  
\centering  
\includegraphics[width=0.85\linewidth]{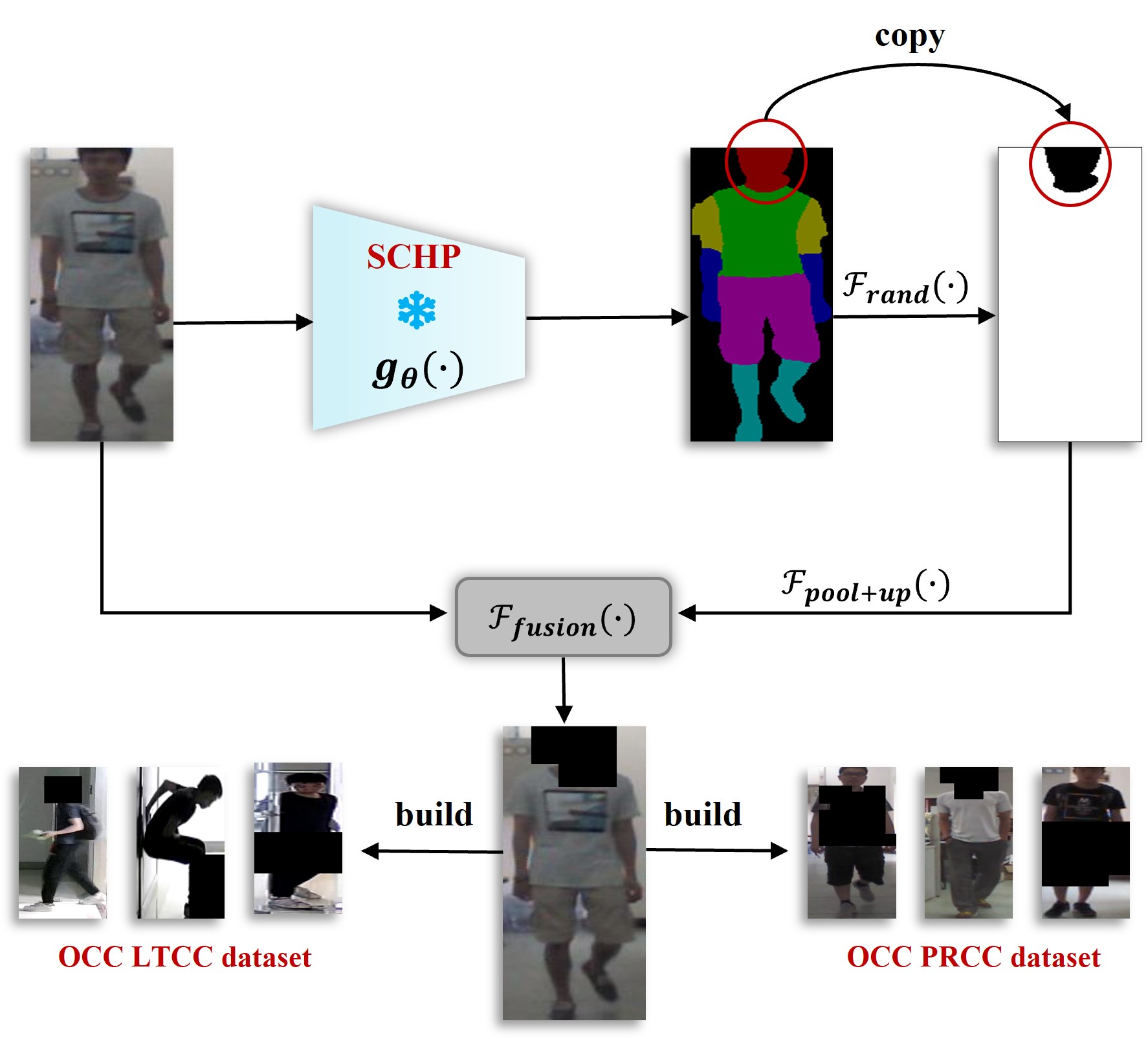}
\caption{Construction details of the occ-PRCC and occ-LTCC datasets.}
\label{fig2}
\end{figure}

As shown in Fig. \ref{fig2}, to facilitate the study of OC\textsuperscript{4}-ReID, we customise two datasets Occ-PRCC and Occ-LTCC based on existing cloth-changing datasets \cite{b12, b13}. Motivated by prior works \cite{b14, b15}, we use black patches to represent occlusions. Specifically, a pre-trained human parser, SCHP \cite{b11}, is employed to locate the key components of pedestrians (i.e., head, torso, upper or lower arms, upper or lower legs). We then randomly select a component to get the occlusion target $f_{\text{mask}}$, and this process is detailed as $f_{\text{mask}} = \mathcal{F}_{\text{rand}} \left( g_{\theta}(x) \right)\label{eq}$, where \( x \) denotes the input image, \( g_{\theta}(\cdot) \) is the processes of SCHP inference, and \( \mathcal{F}_{\text{rand}}(\cdot) \) refers to the random selection process. Note that occlusions in real-world do not exactly match the segmentation regions, therefore we perform a pooling and an up-sampling operation on the segmentation results to obtain more arbitrary occlusion shapes. Finally, we combine the occluded and original image to get the processed image $f_{\text{occ}}$:
\begin{equation}
f_{\text{occ}} = \mathcal{F}_{\text{fusion}} \left( \mathcal{F}_{\text{up}} (\mathcal{F}_{\text{av-pool}}(f_{\text{mask}})), x \right)\label{eq},
\end{equation}
where \( \mathcal{F}_{\text{av-pool}}(\cdot) \) and \( \mathcal{F}_{\text{up}}(\cdot) \) denotes the average pooling and up-sampling operations, and \( \mathcal{F}_{\text{fusion}}(\cdot,\cdot) \) represents the images fusion process.

\section{Method}

\begin{figure*}[t]  
\centering  
\includegraphics[scale=0.5]{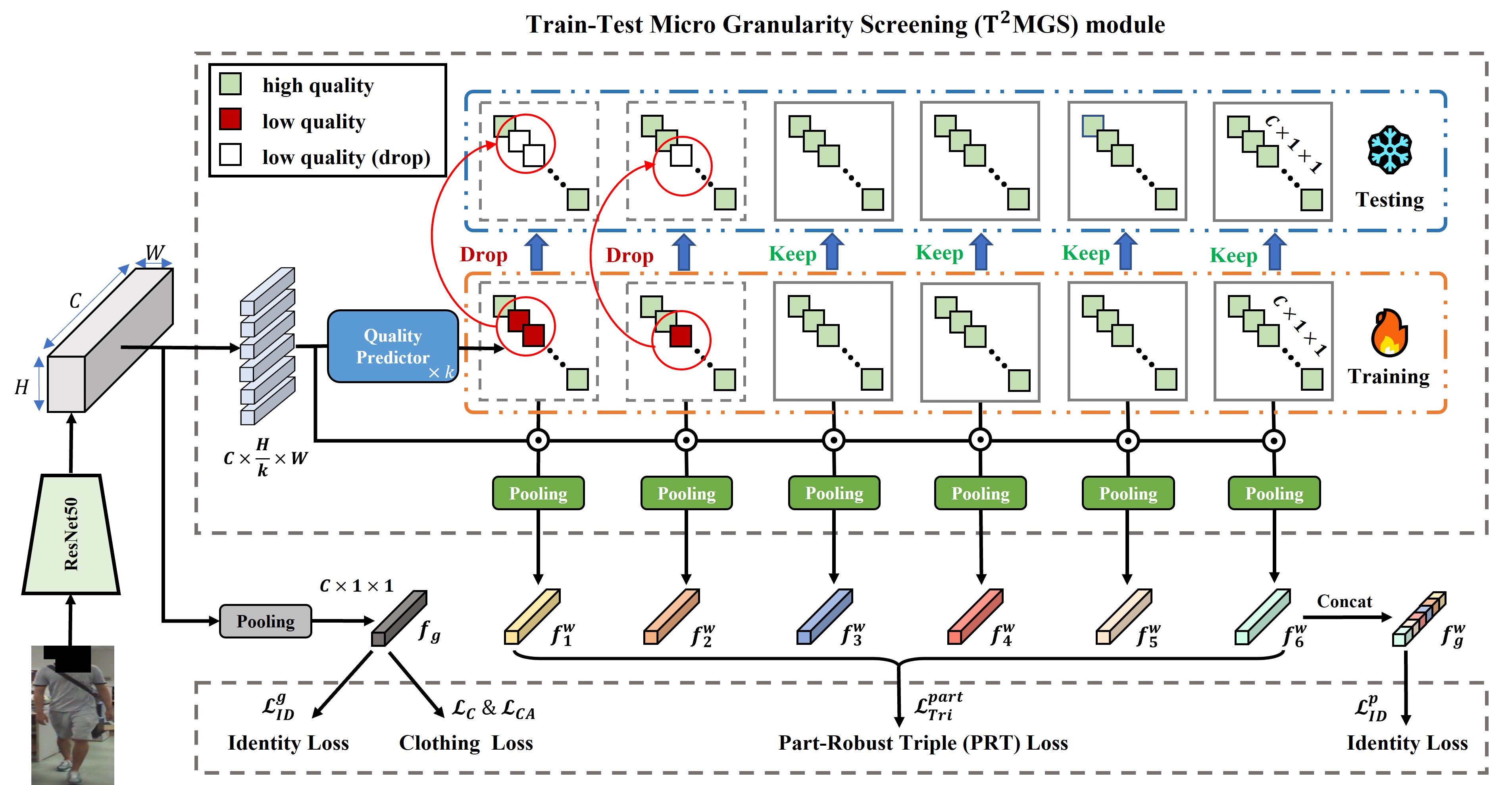}
\caption{We utilize the Train-Test Micro Granularity Screening (T\textsuperscript{2}MGS) module to adaptively discard low-quality features, and the model is supervised by identity loss \cite{b17}, clothing loss \cite{b9}, and the proposed Part-Robust Triple (PRT) loss for learning more discriminative representations.}
\label{fig3}
\end{figure*}

The overall architecture of our method is illustrated in Fig. \ref{fig3}. We first introduce the Train-Test Micro Granularity Screening (T\textsuperscript{2}MGS) module in Section \ref{AA}. Following that, we provide a detailed explanation of the Part-Robust Triplet (PRT) loss in Section \ref{BB}.

\subsection{Train-Test Micro Granularity Screening Module}\label{AA}
\paragraph{Training stage}
Let \( F \in \mathbb{R}^{C \times H \times W} \) denotes the output feature of the backbone, where \( C \), \( H \), and \( W \) are the number of channels, height, and width. We divide \(F\) into \( k \) horizontal partitions denoted as \( f_1, f_2, \dots, f_k \in \mathbb{R}^{C \times \frac{H}{k} \times W} \). For each partition, we maintain a quality predictor, which comprise two convolutional layers, a batch normalization layer, and a sigmoid activation layer. The above process can be formulated as:
\begin{equation}
\mathcal{F}_{Qp}^{i}(\cdot) = \mathcal{F}_{\text{ad-pool}}\left(\sigma ( \mathcal{F}_{\text{BN}}\left(\mathcal{F}_{1 \times 1}\left(\mathcal{F}_{1 \times 1}(\cdot, \theta), \theta\right)\right))\right),
\label{eq}
\end{equation}
\begin{equation}
f_i^{\phi} = \mathcal{F}_{Qp}^{i}(f_i), \quad i \in [1, k],
\label{eq}
\end{equation}
where \(\mathcal{F}_{1 \times 1}(\cdot, \theta)\) denotes the convolution operation with a kernel size of \( 1 \times 1\), and \(\theta \) denotes the process parameters. \(\mathcal{F}_{\text{BN}}(\cdot)\) and \(\sigma (\cdot)\) represent the batch normalization and sigmoid function, respectively. \(\mathcal{F}_{\text{ad-pool}}(\cdot)\) denotes the adaptive average pooling operation. \(f_i^{\phi} \in \mathbb{R}^{C \times 1 \times 1}\) is the feature obtained by the quality predictor, and we dot product it on each partition and followed by a adaptive average pooling to obtain \( f_i^{w} \in \mathbb{R}^{C \times 1 \times 1} \), which is detailed as $f_i^{w}=\mathcal{F}_{\text{ad-pool}}(f_i\cdot f_i^{\phi})$. Subsequently, we concatenate the feature vector\(\ f_i^{w}\) (\(\ i\in \{1, 2, \dots, k\} \)) in the \(H\)-dimension and obtain the global feature\(\ f_g^{w} \). In addition, we also maintain a pooled original feature $f_g$ for the identity and clothing classifiers. The value of \( k \) is set to 6.

\paragraph{Testing stage}

During the testing phase, for the quality prediction result \(f_i^{\phi}(i \in \{1, 2, \dots, k\}) \) of $k$ partitions, we conduct a micro granularity screening in the channel dimension. Specifically, channels with quality scores lower than the threshold \(\lambda \) (set to 0.35 in this paper) are discarded. Through this, our model can adaptively discard occluded features while focusing on discriminative representations. The detailed procedure is outlined in Algorithm 1.

\begin{algorithm}[t]
\small
\caption{Testing Stage}
\begin{algorithmic}[1]
\IF{not training}
    \FOR{each tensor in $f_i^{\phi}$}
        \FOR{each channel in tensor}
            \IF{channel $\textless$ $\lambda$}
                \STATE Set channel to 0
            \ENDIF
        \ENDFOR   
    \ENDFOR
    \FOR{i = 0 to $k$}
        \STATE $f_i^{\phi} = \mathcal{F}_{\text{ad-pool}}(f_i^{\phi} \cdot f_i)$ 
    \ENDFOR
    \STATE $f_{tmp} \gets$ Concatenate tensors in $f_i^{\phi}$ along dimension $H$
    \STATE $f_g^{w} = \mathcal{F}_{\text{ad-pool}}(f_{tmp})$ 
\ENDIF
\RETURN $f_g^{w}$
\end{algorithmic}
\end{algorithm}

\subsection{Part-Robust Triplet Loss}\label{BB}
In part-based ReID methods \cite{b18, b19, b20, b21}, identity-related triplet loss \cite{b16} is directly used to supervise partial features. However, these approaches encounter limitations in the OC\textsuperscript{4}-ReID scenario: (i) Generally, pedestrian clothing is more likely to act as a discriminative feature and promote the ReID task. However, clothing features are interference information in OC\textsuperscript{4}-ReID, supervising the clothing region with identity-related loss impairs model performance. (ii) Occluded regions mainly contain blurred and low-quality features with little identity information. Using identity-related loss for supervision introduces misleading information. To address these issues, we propose the Part-Robust Triplet (PRT) Loss, which considers the average distances of all partial features and focuses on mining the overall discriminative representations:
\begin{equation}
\mathcal{D}(x, y) = \frac{1}{k} \sum_{i=1}^{k} \mathcal{F}_{\text{dis}}^{\text{eucl}}(f_i^{x}, f_i^{y}),\label{eq}
\end{equation}
where \( \mathcal{F}_{dis}^{eucl}(\cdot) \) denotes the Euclidean distance. For the anchor sample $a$, we compute the hardest positive distance \( \mathcal{D}_{(a,p)} \) and the hardest negative distance \( \mathcal{D}_{(a,n)} \), similar to \cite{b16}. Finally, the part-robust triplet loss can be expressed as follows:
\begin{equation}
\mathcal{L}_{\text{Tri}}^{\text{part}} = \sum \mathcal{F}_{\text{max}}\left(\mathcal{D}_{(a,p)} - \mathcal{D}_{(a,n)} + M, 0\right),\label{eq}
\end{equation}
where \( \mathcal{F}_{\text{max}}(\cdot, \cdot) \) returns the maximum value, and \( M \) is the margin in triplet loss, which is set to 0.3 in this paper. For global features \( f_g \) and \( f_g^{w} \), we utilize cross-entropy loss \cite{b17} as the identity loss to supervise, which are denoted as \( \mathcal{L}_{ID}^{g} \) and \( \mathcal{L}_{ID}^{p} \). Additionally, similar to \cite{b9}, we introduce a clothing-based adversarial loss \( \mathcal{L}_{CA} \) that collaborates with the clothing classification loss \( \mathcal{L}_{C} \) \cite{b17} to learn the cloth-unrelated features. The final total loss is calculated as follows:
\begin{equation}
\mathcal{L} = \mathcal{L}_{\text{Tri}}^{\text{part}} + \mathcal{L}_{ID}^{p} + \mathcal{L}_{ID}^{g} + \mathcal{L}_{C} + \mathcal{L}_{CA}.\label{eq}
\end{equation}

\section{Experiments}

\subsection{Datasets and Evaluation Metrics}
We evaluate the performance of our method on two CC-ReID benchmark datasets PRCC \cite{b12} and LTCC \cite{b13}, as well as the proposed Occ-PRCC and Occ-LTCC datasets, under the cloth-changing configuration. 
PRCC \cite{b12} is a large CC-ReID dataset consists of 33,698 pedestrian images from 221 identities, and each person wears the same clothes in cameras 1 and 2, and different clothes in cameras 1 and 3. LTCC \cite{b13} contains 17,119 images from 152 identities and 478 outfits, which is captured with 12 different views. Following the previous studies \cite{b9, b10}, this paper use Rank@1 and mean Average Precision (mAP) for evaluation.

\subsection{Implementation Details}
We implement our method with PyTorch and train the model on two NVIDIA RTX 4090 with 24 Gigabyte memory. ResNet-50 \cite{b11} pre-trained on ImageNet is selected as the backbone. All the input images are resized to $384\times192$. Random horizontal flipping, cropping and erasing are utilized for data augmentation. The batch size is set to 64, and initial learning 
rate is set to $3.5 \times 10^{-4}$, which decays at 30th and 50th epochs with a decay factor of 0.1 (120 epochs in total).

\begin{table*}[!t]
\caption{Comparison with state-of-the-art methods on the standard CC-ReID datasets and OC\textsuperscript{4}-ReID datasets.}
\centering
\scriptsize 
\setlength{\tabcolsep}{4pt} 
\renewcommand{\arraystretch}{1.2} 
\begin{tabular}{l|cc|cc|cc|cc}
\hline \hline
\multicolumn{1}{c|}{\multirow{2}{*}{Method}} & \multicolumn{2}{c|}{Occ-PRCC}   & \multicolumn{2}{c|}{Occ-LTCC}   & \multicolumn{2}{c|}{PRCC}       & \multicolumn{2}{c}{LTCC}        \\ \cline{2-9} 
\multicolumn{1}{c|}{}                        & Rank@1         & mAP            & Rank@1         & mAP            & Rank@1         & mAP            & Rank@1         & mAP            \\ \hline
HACNN \cite{b23}                                & 10.12          & 7.71           & 12.73          & 5.92           & 21.86          & 17.61          & 21.61          & 9.32           \\
PCB \cite{b24}                                  & 30.64          & 28.73          & 14.94          & 7.41           & 41.87          & 38.71          & 23.52          & 10.12          \\
IANet \cite{b25}                                & 35.51          & 35.51          & 16.54          & 8.64           & 46.32          & 45.92          & 25.21          & 12.61          \\
RCSANet \cite{b26}                              & 41.95          & 36.93          & 22.56              & 9.04              & 50.24          & 48.67          & 29.61              & 13.87              \\
CAL \cite{b9}                                   & 45.34          & 43.21          & 29.94          & 9.23           & 55.21          & 55.83          & 34.41          & 16.12          \\
AIM \cite{b10}                                 & 47.10          & 44.98          & 31.51          & 11.43          & 57.90          & 58.30          & 40.60          & 19.10          \\
\textbf{Ours*}                                & \textbf{58.81} & \textbf{56.52} & \textbf{40.73} & \textbf{19.89} & \textbf{62.28} & \textbf{62.12} & \textbf{43.84} & \textbf{22.43} \\ \hline \hline
\end{tabular}
\label{tab1}
\end{table*}

\subsection{Compare with State-of-the-art Methods}

We compare the proposed model with three traditional ReID architectures (i.e. HACNN \cite{b23}, PCB \cite{b24}, and IANet \cite{b25}) and three classical open-source CC-ReID methods (i.e. RCSANet \cite{b26}, CAL \cite{b9}, AIM \cite{b10}) in Table \ref{tab1}.

\textbf{Result on Occ-PRCC and Occ-LTCC:} As shown in Table \ref{tab1}, our model outperforms existing state-of-the-art methods on the OC\textsuperscript{4}-ReID datasets. Taking the Occ-PRCC dataset as an example, our model achieves 58.81\% Rank-1 and 56.52\% mAP, which is 11.71\% and 11.54\% higher than the second-best method. Similarly, we also achieve advanced performance on the Occ-LTCC dataset, with Rank-1 and mAP improving by 9.22\% and 8.46\% compared to the second-best method. The experimental results demonstrate the superior performance of our method in handling the OC\textsuperscript{4}-ReID task, because the following two reasons: (i) By conducting screening mechanisms on local and channel features, T\textsuperscript{2}MGS effectively suppresses low-quality features while further mining fine-grained discriminative parts. (ii) Using PRT loss to supervise partial features, this key design motivates the model to focus on truly discriminative features at each training step, which in turn mitigates the effect of occlusions and clothing features.

\textbf{Result on PRCC and LTCC:} In addition, we compare our method with current state-of-the-art methods on standard CC-ReID datasets: PRCC \cite{b12} and LTCC \cite{b13}. On the LTCC dataset, the Rank-1 and mAP of our method under the cloth-changing scenario are 43.84\% and 22.43\%, outperforming the second-best method by 3.24\% and 3.33\%, respectively. Similarly, our method also achieve the optimal performance on PRCC dataset. The experimental results demonstrate that our method can also achieve excellent performance in a separate cloth-changing scenario.

\subsection{Ablation Study of The Proposed Method}

\begin{table}[!t]
\centering
\caption{Ablation Experiments of Our Method on Occ-PRCC Dataset.}
\resizebox{\linewidth}{!}{
\begin{tabular}{cccc|cc}
\hline \hline
\multirow{2}{*}{Index} & \multirow{2}{*}{T\textsuperscript{2}MGS} & \multirow{2}{*}{PRT loss} & \multirow{2}{*}{Triplet loss} & \multicolumn{2}{c}{Occ-PRCC}    \\ \cline{5-6} 
                       &                                   &                                           &                               & Rank@1         & mAP            \\ \hline
0(baseline)                      &                                   &                                          &                               & 45.34          & 43.21          \\
1                      & \checkmark                                 &                                           &                               & 54.73          & 54.65          \\
2                      & \checkmark                                 &                                           & \checkmark                             & 52.12          & 52.11          \\
\textbf{3(Ours*)}       & \textbf{\checkmark}                        & \textbf{\checkmark}                                & \textbf{}                     & \textbf{58.81} & \textbf{56.52} \\ \hline \hline
\end{tabular}
}
\label{tab}
\end{table}

\begin{figure}[!t]  
\centering  
\includegraphics[width=\linewidth]{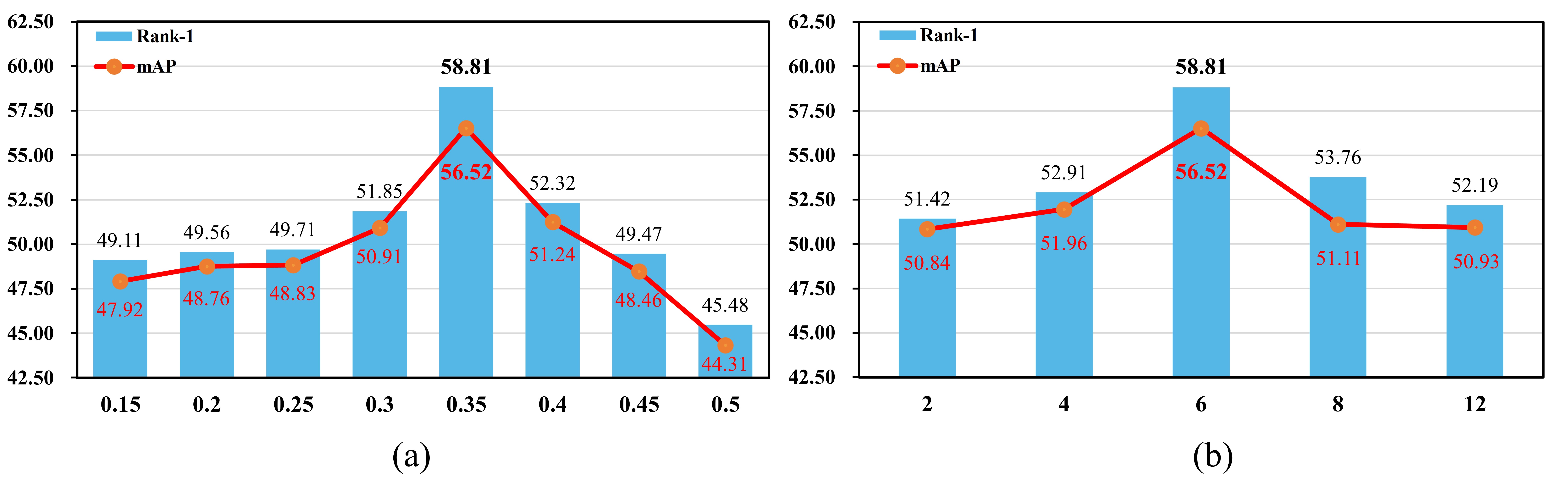}
\caption{Parameter-sensitivity experiments. (a) and (b) evaluate our method with different values of \( \lambda \) and \( k \) on Occ-PRCC dataset.}
\label{fig4}
\end{figure}

Table \ref{tab} reports the ablation experimental results of our method in Occ-PRCC. Our baseline achieves 45.34\% Rank@1 and 43.21\% mAP accuracy. When T\textsuperscript{2}MGS is introduced for occlusion mitigation (Index 1), the Rank@1 and mAP are improved by 9.39\% and 11.44\%, respectively. Then, we use classic triplet loss \cite{b16} for supervising partial features (Index 2), which reduce the Rank@1 and mAP by 2.61\% and 2.54\%. The results show that acting triplet loss directly on partial features inevitably damages the feature extraction process and introduces interference information. Finally, when replacing triplet loss with the proposed PRT loss (Index 3), our method achieves optimal performance with Rank@1 and mAP of 58.81\% and 56.52\%.

\subsection{Sensitivity Analysis of Hyper-Parameter Setting}

To explore the optimal value of threshold \( \lambda \) and partition number \( k \), we conduct parameter-sensitivity experiments on the Occ-PRCC dataset. We first set \( \lambda \) and \( k \) to 0.15 and 2, then increase them by 0.05 and 2, respectively. As shown in Fig. \ref{fig4} (a), the model's performance reduces when the threshold \(\lambda \) once exceeds 0.35, which indicates that visible features are removed along with occluded features. In addition, as illustrated in Fig. \ref{fig4} 
(b), the model achieves best performance when the number of partitions \( k \) is set to 6.

\begin{figure}[!t]  
\centering  
\includegraphics[width=\linewidth]{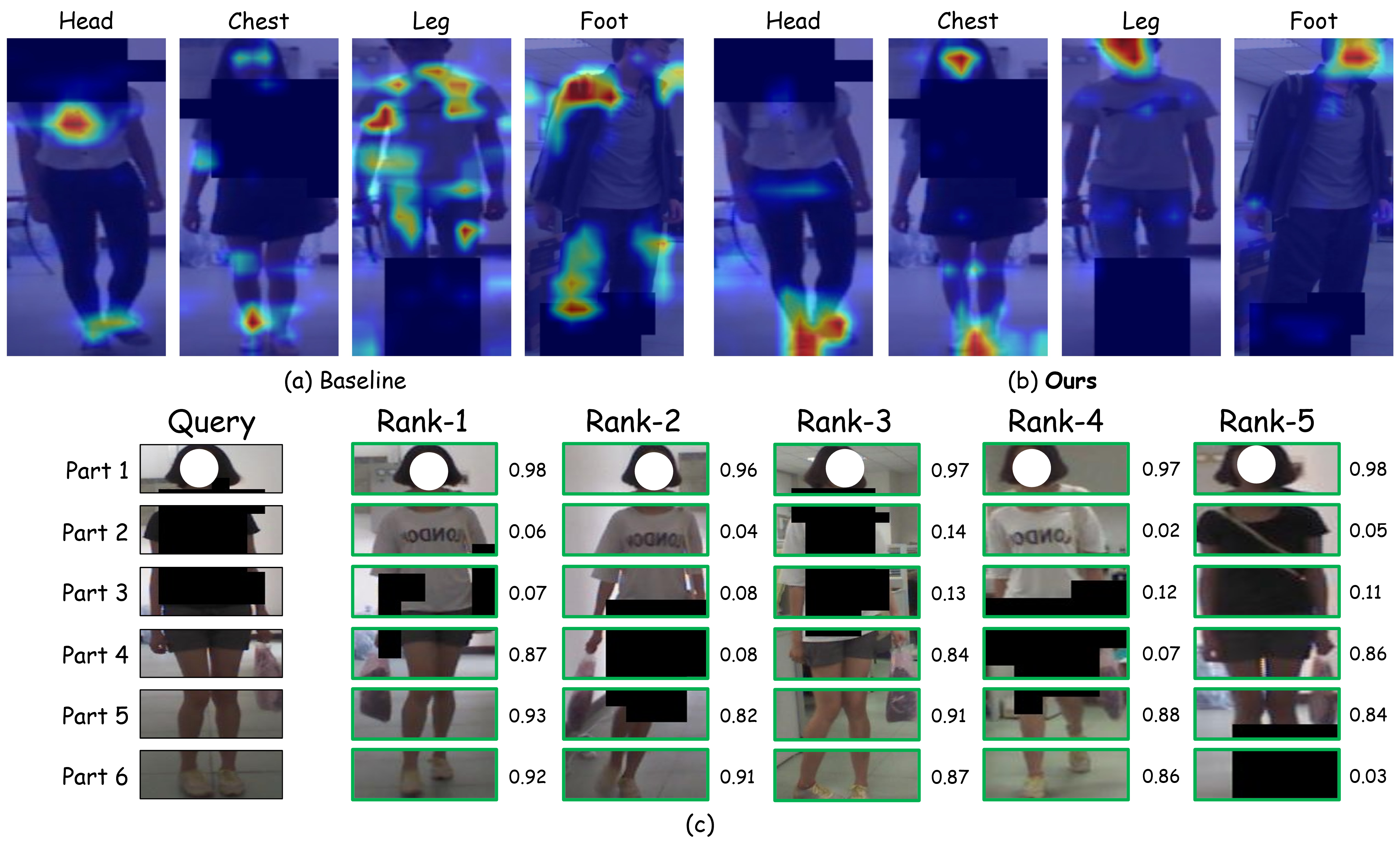}
\caption{(a) and (b) are the Grad-CAM visualization of the baseline and our method on PRCC datasets. (c) visualizes retrieval results and part-based similarities/distances.}
\label{fig5}
\end{figure}

\subsection{Visualization}
\begin{figure}[!t]  
\centering  
\includegraphics[width=\linewidth]{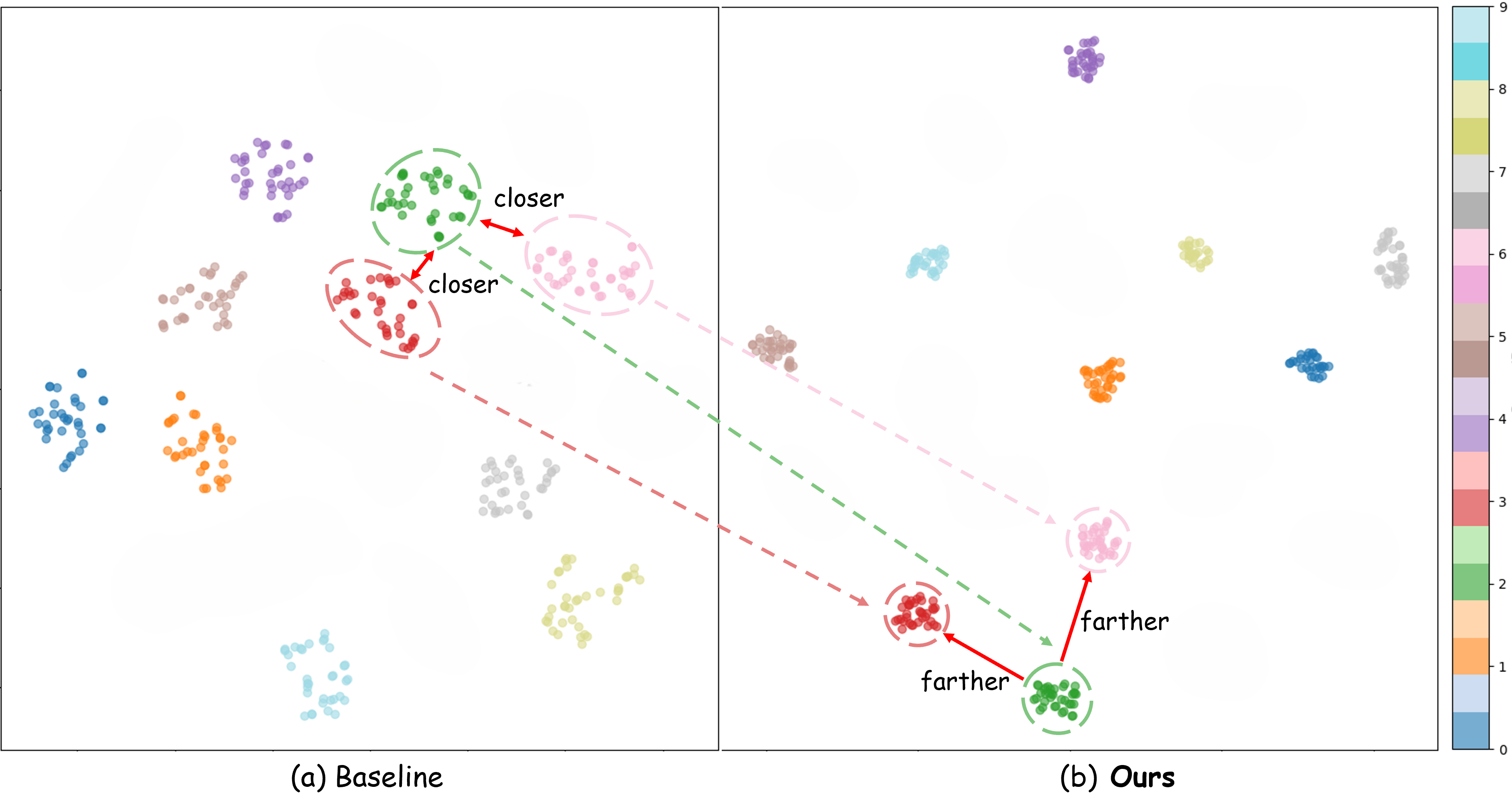}
\caption{t-SNE feature visualization of baseline and our method on PRCC dataset. Different colors represent different identities.}
\label{fig6}
\end{figure}

We conduct visualization experiments to compare OC\textsuperscript{4}-ReID with the baseline. In Fig.\ref{fig5} (a) and (b), we present the Grad-CAM \cite{b27} visualization results, showing that our method focuses more effectively on discriminative pedestrian appearance, whereas the baseline is frequently influenced by identity-independent fragmented information. Additionally, Fig.\ref{fig5} (c) visualizes part-based similarities. The results demonstrate that the model effectively assigns low scores to occluded and cloth-related areas while allocating high scores to cloth-unrelated discriminative body parts, highlighting the model's ability to handle occlusion and cloth-changing. As demonstrated in Fig. \ref{fig6}, we utilize t-SNE \cite{b28} to visualize the feature distribution of random ten identities of baseline and our method. It can be observed that our method encourages more compact intra-class representations while increasing inter-class discrimination, which further validates the effectiveness of our proposed approach.

\section{Conclusion}
In this paper, we propose the task of cloth-changing person re-identification under occlusion conditions, termed OC\textsuperscript{4}-ReID, for the first time. To support this new task, we construct two specialized datasets, Occ-PRCC and Occ-LTCC. To address the challenges presented in this task, we design a new benchmark, which includes a Train-Test Micro Granularity Screening (T\textsuperscript{2}MGS) module. This module effectively mitigates occlusions while facilitating the extraction of fine-grained cloth-unrelated features. Furthermore, we propose a Part-Robust Triplet (PRT) loss, which supervises the partial features learning properly. Experimental results on both CC-ReID and OC\textsuperscript{4}-ReID datasets demonstrate the effectiveness of our proposed method.

\end{document}